\documentclass[10pt,twocolumn,letterpaper]{article}

\usepackage{cvpr}              

\usepackage[utf8]{inputenc}
\usepackage[T1]{fontenc}

\usepackage{color,xcolor}
\usepackage{epsfig}
\usepackage{graphicx}

\usepackage{adjustbox}
\usepackage{array}
\usepackage{booktabs}
\usepackage{colortbl}
\usepackage{float,wrapfig}
\usepackage{hhline}
\usepackage{multirow}
\usepackage{subcaption} %
\usepackage[font=small]{caption}

\usepackage{amsmath,amsfonts,amsthm,amssymb}
\usepackage{bm}
\usepackage{nicefrac}
\usepackage{microtype}

\usepackage{changepage}
\usepackage{extramarks}
\usepackage{fancyhdr}
\usepackage{lastpage}
\usepackage{setspace}
\usepackage{soul}
\usepackage{xspace}
\usepackage{indentfirst}
\usepackage{pifont}
\usepackage{cuted}
\usepackage{wrapfig}

\usepackage[pagebackref,breaklinks,colorlinks,urlcolor=blue,citecolor=cvprblue]{hyperref}
\usepackage{url}

\usepackage{algorithm, algorithmic}
\usepackage{enumitem}

\usepackage{wasysym}
\usepackage{duckuments}
\usepackage{pifont}
\newcolumntype{L}[1]{>{\raggedright\let\newline\\\arraybackslash\hspace{0pt}}m{#1}}
\newcolumntype{C}[1]{>{\centering\let\newline\\\arraybackslash\hspace{0pt}}m{#1}}
\newcolumntype{R}[1]{>{\raggedleft\let\newline\\\arraybackslash\hspace{0pt}}m{#1}}

\newcommand{\sect}[1]{Section~\ref{sect:#1}}

\newcommand{\fig}[1]{Figure~\ref{fig:#1}}

\newcommand{\tab}[1]{Table~\ref{tab:#1}}

\newcommand{\fid}{Fr\'echet Inception Distance\xspace}
\newcommand{\lblfig}[1]{\label{fig:#1}}
\newcommand{\lblsect}[1]{\label{sect:#1}}

\newcommand{\lbltab}[1]{\label{tab:#1}}

\newcommand{\ignorethis}[1]{}

\def\naive{na\"{\i}ve\xspace}
\def\Naive{Na\"{\i}ve\xspace}
\def\naively{na\"{\i}vely\xspace}
\def\Naively{Na\"{\i}vely\xspace}

\makeatletter
\DeclareRobustCommand\onedot{\futurelet\@let@token\@onedot}
\def\@onedot{\ifx\@let@token.\else.\null\fi\xspace}

\def\eg{\emph{e.g}\onedot} 
\def\ie{\emph{i.e}\onedot}

\def\etal{\emph{et al}\onedot}
\makeatother
\definecolor{citecolor}{rgb}{34,139,34}
\definecolor{mydarkblue}{rgb}{0,0.08,1}
\definecolor{mydarkgreen}{rgb}{0.02,0.6,0.02}
\definecolor{mydarkred}{rgb}{0.8,0.02,0.02}
\definecolor{mydarkorange}{rgb}{0.40,0.2,0.02}
\definecolor{mypurple}{RGB}{111,0,255}
\definecolor{myred}{rgb}{1.0,0.0,0.0}
\definecolor{mygold}{rgb}{0.75,0.6,0.12}
\definecolor{mydarkgray}{rgb}{0.66,0.66,0.66}

\newcommand{\myparagraph}[1]{\vspace{1pt}\noindent\textbf{#1}}

\def\multirowcenter{-0.5\dimexpr \aboverulesep + \belowrulesep + \cmidrulewidth}

\def\parallelism{displaced patch parallelism\xspace}
\def\Parallelism{Displaced patch parallelism\xspace}
\def\method{DistriFusion\xspace}

\definecolor{cvprblue}{rgb}{0.21,0.49,0.74}

\def\paperID{4189} 
\def\confName{CVPR}
\def\confYear{2024}

\title{DistriFusion: Distributed Parallel Inference for High-Resolution Diffusion Models}

\author{Muyang Li$^{1}\thanks{ indicates equal contributions.}$
\qquad
Tianle Cai$^{2}\footnotemark[1]$
\quad
Jiaxin Cao$^{3}$
\quad
Qinsheng Zhang$^{4}$
\quad
Han Cai$^{1}$
\\
Junjie Bai$^{3}$
\quad
Yangqing Jia$^{3}$
\quad
Ming-Yu Liu$^{4}$
\quad
Kai Li$^{2}$
\quad
Song Han$^{1,4}$ \\
\\
$^1$MIT \quad $^2$Princeton \quad $^3$Lepton AI \quad $^4$NVIDIA \\
\url{https://github.com/mit-han-lab/distrifuser}
}

\begin{document}

\maketitle
\begin{strip}
    \centering
    \vspace{-45pt}
    \includegraphics[width=\linewidth]{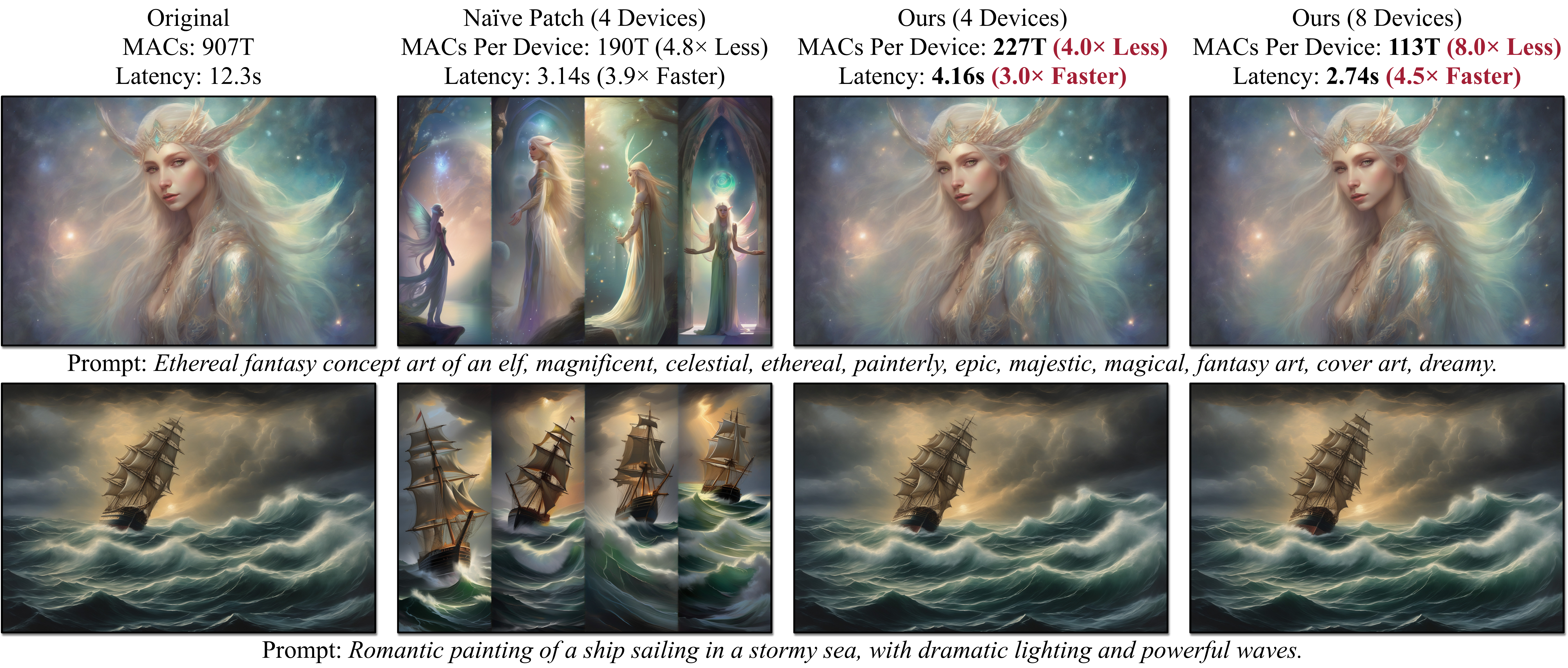}
    \vspace{-20pt}
    \captionof{figure}{
        \looseness=-1 \looseness=-1 We introduce \method, a training-free algorithm to harness multiple GPUs to accelerate diffusion model inference without sacrificing image quality. \Naive Patch (\fig{idea}(b)) suffers from the fragmentation issue due to the lack of patch interaction. Our DistriFusion removes artifacts and avoids the communication overhead by reusing the features from the previous steps. Setting: SDXL with 50-step Euler sampler, $1280\times1920$ resolution. Latency is measured on A100s.
    }
    \vspace{-5pt}
    \lblfig{teaser}
\end{strip}
\begin{abstract}
    Diffusion models have achieved great success in synthesizing high-quality images. However, generating high-resolution images with diffusion models is still challenging due to the enormous computational costs, resulting in a prohibitive latency for interactive applications. In this paper, we propose \method to tackle this problem by leveraging parallelism across multiple GPUs. Our method splits the model input into multiple patches and assigns each patch to a GPU. However, \naively implementing such an algorithm breaks the interaction between patches and loses fidelity, while incorporating such an interaction will incur tremendous communication overhead. To overcome this dilemma, we observe the high similarity between the input from adjacent diffusion steps and propose \parallelism, which takes advantage of the sequential nature of the diffusion process by reusing the pre-computed feature maps from the previous timestep to provide context for the current step. Therefore, our method supports asynchronous communication, which can be pipelined by computation. Extensive experiments show that our method can be applied to recent Stable Diffusion XL with no quality degradation and achieve up to a 6.1$\times$ speedup on eight A100 GPUs compared to one.
    \vspace{-20pt}
\end{abstract}

\section{Introduction}
\looseness=-1
The advent of AI-generated content (AIGC) represents a seismic shift in technological innovation. Tools like \href{https://www.adobe.com/sensei/generative-ai/firefly.html}{Adobe Firefly}, \href{https://www.midjourney.com/home?callbackUrl=%2Fexplore}{Midjourney} and recent \href{https://openai.com/sora}{Sora} showcase astonishing capabilities, producing compelling imagery and designs from simple text prompts. These achievements are notably supported by the progression in diffusion models~\cite{ho2020denoising,sohl2015deep}. The emergence of large text-to-image models, including Stable Diffusion~\cite{rombach2022high}, Imgen~\cite{saharia2022photorealistic}, eDiff-I~\cite{balaji2022ediffi}, DALL$\cdot$E~\cite{ramesh2021zero,ramesh2022hierarchical,betker2023improving} and Emu~\cite{dai2023emu}, further expands the horizons of AI creativity. Trained on diverse open-web data, these models can generate photorealistic images from text descriptions alone. Such technological revolution unlocks numerous synthesis and editing applications for images and videos, placing new demands on responsiveness: by \textit{interactively} guiding and refining the model output, users can achieve more personalized and precise results. Nonetheless, a critical challenge remains --  high resolution leading to large computation. For example, the original Stable Diffusion~\cite{rombach2022high} is limited to generating $512\times512$ images. Later, SDXL~\cite{podell2023sdxl} expands the capabilities to $1024\times1024$ images. More recently, Sora further pushes the boundaries by enabling video generation at $1080\times1920$ resolution. Despite these advancements, the increased latency of generating high-resolution images presents a tremendous barrier to real-time applications.

Recent efforts to accelerate diffusion model inference have mainly focused on two approaches: reducing sampling steps~\cite{salimans2021progressive,song2020denoising,meng2022distillation,kong2021fast,xiao2022DDGAN,lu2022dpm,lu2022dpm++,zhang2022fast} and optimizing neural network inference~\cite{li2022efficient,li2023q,li2023snapfusion}. As computational resources grow rapidly, leveraging multiple GPUs to speed up inference is appealing. For example, in natural language processing (NLP), large language models have successfully harnessed tensor parallelism across GPUs, significantly reducing latency.
However, for diffusion models, multiple GPUs are usually only used for batch inference. When generating a single image, typically only one GPU is involved (\fig{idea}(a)). Techniques like tensor parallelism are less suitable for diffusion models due to the large activation size, as communication costs outweigh savings from distributed computation. Thus, even when multiple GPUs are available, they cannot be effectively exploited to further accelerate single-image generation. This motivates the development of a method that can utilize multiple GPUs to speed up single-image generation with diffusion models.

\looseness=-1
A \naive approach would be to divide the image into several patches, assigning each patch to a different device for generation, as illustrated in \fig{idea}(b). This method allows for independent and parallel operations across devices. However, it suffers from a clearly visible seam at the boundaries of each patch due to the absence of interaction between the individual patches. However, introducing interactions among patches to address this issue would incur excessive synchronization costs again, offsetting the benefits of parallel processing.

\begin{figure}[t]
    \centering
    \includegraphics[width=\linewidth]{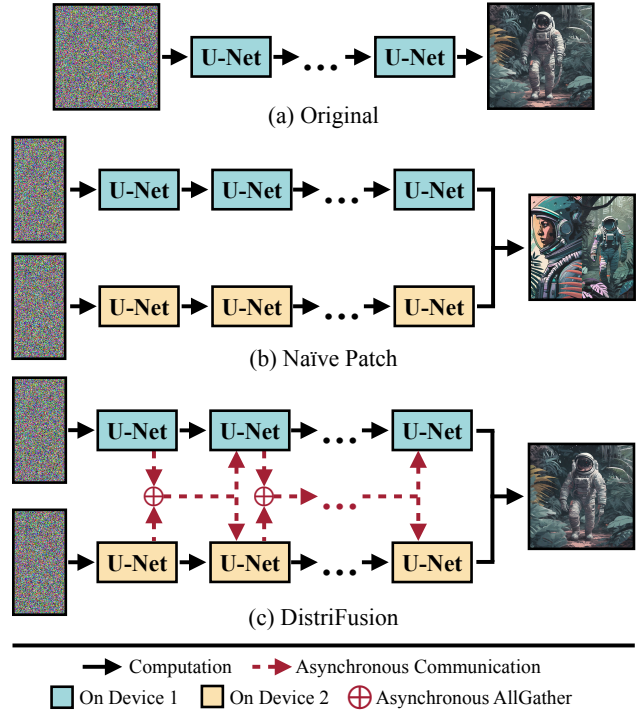}
    \vspace{-15pt}
	\caption{
        (a) Original diffusion model running on a single device. (b) \Naively splitting the image into 2 patches across 2 GPUs has an evident seam at the boundary due to the absence of interaction across patches. (c) \method employs synchronous communication for patch interaction at the first step. After that, we reuse the activations from the previous step via asynchronous communication. In this way, the communication overhead can be hidden into the computation pipeline.
	}
    \lblfig{idea}
    \vspace{-15pt}
\end{figure}

\looseness=-1
In this work, we present \emph{\method}, a method that enables running diffusion models across multiple devices in parallel to reduce the latency of single-sample generation without hurting image quality. As depicted in \fig{idea}(c), our approach is also based on patch parallelism, which divides the image into multiple patches, each assigned to a different device. Our key observation is that the inputs across adjacent denoising steps in diffusion models are similar. Therefore, we adopt synchronous communication solely for the first step. For the subsequent steps, we reuse the pre-computed activations from the \textit{previous step} to provide global context and patch interactions for the \textit{current step}. We further co-design an inference framework to implement our algorithm. Specifically, our framework effectively hides the communication overhead within the computation via asynchronous communication. It also sparsely runs the convolutional and attention layers exclusively on the assigned regions, thereby proportionally reducing per-device computation. Our method, distinct from data, tensor, or pipeline parallelism, introduces a new parallelization opportunity: \emph{\parallelism}.

\looseness=-1
\method only requires off-the-shelf pre-trained diffusion models and is applicable to a majority of few-step samplers. We benchmark it on a subset of COCO Captions~\cite{chen2015microsoft}. Without loss of visual fidelity, it mirrors the performance of the original Stable Diffusion XL (SDXL)~\cite{podell2023sdxl} while reducing the computation\footnote{Following previous works, we measure the computational cost with the number of Multiply-Accumulate operations (MACs). 1 MAC=2 FLOPs.} proportionally to the number of used devices. Furthermore, our framework also reduces the latency of SDXL U-Net for generating a single image by up to 1.8$\times$, 3.4$\times$ and 6.1$\times$ with 2, 4, and 8 A100 GPUs, respectively. When combined with batch splitting for classifier-free guidance~\cite{ho2021classifier},  we achieve in total 3.6$\times$ and 6.6$\times$ speedups using 4 and 8 A100 GPUs for $3840\times3840$ images, respectively. See \fig{teaser} for some examples of our method.

\section{Related Work}
\looseness=-1
\myparagraph{Diffusion models.} 
Diffusion models have significantly transformed the landscape of content generation~\citep{ho2020denoising,podell2023sdxl,nichol2022glide,balaji2022ediffi}. At its core, these models synthesize content through an iterative denoising process. Although this iterative approach yields unprecedented capabilities for content generation, it requires substantially more computational resources and results in slower generative speed. This issue intensifies with the synthesis of high-dimensional data, such as high-resolution~\cite{hoogeboom2023simple,gu2023matryoshka} or $360^\circ$ images~\cite{zhange2023diffcollage}. Researchers have investigated various perspectives to accelerate the diffusion model. The first line lies in designing more efficient denoising processes. Rombach \etal~\citep{rombach2022high} and Vahdat~\etal~\citep{vahdat2021score} propose to compress high-resolution images into low-resolution latent representations and learn diffusion model in latent space. Another line lies in improving sampling via designing efficient training-free sampling algorithms. A large category of works along this line is built upon the connection between diffusion models and differential equations~\citep{song2020score}, and leverage a well-established exponential integrator~\citep{zhang2022fast,zhang2022gddim,lu2022dpm} to reduce sampling steps while maintaining numerical accuracy. The third strategy involves distilling faster generative models from pre-trained diffusion models. Despite significant progress made in this area, a quality gap persists between these expedited generators and diffusion models~\cite{kim2021diffusionclip, salimans2021progressive, meng2022distillation}. In addition to the above schemes, some works investigate how to optimize the neural inference for diffusion models~\cite{li2022efficient,li2023q,li2023snapfusion}. In this work, we explore a new paradigm for accelerating diffusion by leveraging parallelism to the neural network on multiple devices. 

\myparagraph{Parallelism.} 
Existing work has explored various parallelism strategies to accelerate the training and inference of large language models (LLMs), including data, pipeline~\cite{huang2019gpipe,narayanan2019pipedream,li2021terapipe}, tensor~\cite{jia2019beyond,narayanan2021efficient,xu2021gspmd,yuan2021oneflow,zheng2022alpa}, and zero-redundancy parallelism~\cite{rajbhandari2019zero,rasley2020deepspeed,DBLP:conf/usenix/0015RARYZ0H21,zhao2023pytorch}. Tensor parallelism, in particular, has been widely adopted for accelerating LLMs~\cite{li2023alpaserve}, which are characterized by their substantial model sizes, whereas their activation sizes are relatively small. In such scenarios, the communication overhead introduced by tensor parallelism is relatively minor compared to the substantial latency benefits brought by increased memory bandwidth. However, the situation differs for diffusion models, which are generally smaller than LLMs but are often bottlenecked by the large activation size due to the spatial dimensions, especially when generating high-resolution content. The communication overhead from tensor parallelism becomes a significant factor, overshadowing the actual computation time. As a result, only data parallelism has been used thus far for diffusion model serving, which provides no latency improvements. The only exception is ParaDiGMS~\cite{shih2023paradigms}, which uses Picard iteration to run multiple steps in parallel. However, this sampler tends to waste much computation, and the generated results exhibit significant deviation from the original diffusion model. Our method is based on patch parallelism, which distributes the computation across multiple devices by splitting the input into small patches. Compared to tensor parallelism, such a scheme has superior independence and reduced communication demands. Additionally, it favors the use of \texttt{AllGather} over \texttt{AllReduce} for data interaction, significantly lowering overhead (see \sect{Communication cost.} for the full comparisons). Drawing inspiration from the success of asynchronous communication in parallel computing~\cite{valiant1990bridging}, we further reuse the features from the previous step as context for current step to overlap communication and computation, called \emph{\parallelism}. This represents the first parallelism strategy tailored to the sequential characteristics of diffusion models while avoiding the heavy communication costs of traditional techniques like tensor parallelism.

\myparagraph{Sparse computation.} 
Sparse computation has been extensively researched in various domains, including weight~\cite{han2015learning,li2016pruning,liu2015sparse,jaderberg2014speeding}, input~\cite{tang2022torchsparse,riegler2017octnet,tangandyang2023torchsparse} and activation~\cite{ren2018sbnet,judd2017cnvlutin2,shi2017speeding,dong2017more,pan2018recurrent,li2022efficient,li2017not,ren2018sbnet}. In the activation domain, to facilitate on-hardware speedups, several studies propose to use structured sparsity. SBNet~\cite{ren2018sbnet} employs a spatial mask to sparsify activations for accelerating 3D object detection. This mask can be derived either from prior problem knowledge or an auxiliary network. In the context of image generation, SIGE~\cite{li2022efficient} leverages the highly structured sparsity of user edits, selectively performing computation at the edited regions to speed up GANs~\cite{goodfellow2014generative} and diffusion models. MCUNetV2\cite{lin2021mcunetv2} adopts a patch-based inference to reduce memory usage for image classification and detection. In our work, we also partition the input into patches, each processed by a different device. However, we focus on reducing the latency by parallelism for image generation instead. Each device will solely process the assigned regions to reduce the per-device computation.
\section{Background}
\looseness=-1
To generate a high-quality image, a diffusion model often trains a noise-prediction neural model (\eg, U-Net~\cite{ronneberger2015u}) $\epsilon_\theta$. Starting from pure Gaussian noise $\mathbf x_T \sim \mathcal N(\mathbf 0, \mathbf I)$, it involves tens to hundreds of iterative denoising steps to get the final clean image $\mathbf x_0$, where $T$ is the total number of steps. Specifically, given the noisy image $\mathbf x_t$ at time step $t$, the model $\epsilon_\theta$ takes $\mathbf x_t$, $t$ and an additional condition $c$ (\eg, text) as inputs to predict the corresponding noise $\epsilon_t$ within $\mathbf x_t$. At each denoising step, $\mathbf x_{t-1}$ can be derived from the following equation:
\begin{equation}
    \mathbf x_{t-1} = \text{Update}(\mathbf x_{t}, t, \epsilon_t), \quad \epsilon_t = \epsilon_\theta(\mathbf x_t, t, c).
\end{equation}
Here, `$\text{Update}$' refers to a sampler-specific function that typically includes element-wise additions and multiplications. Therefore, the primary source of latency in this process is the forward passes through model $\epsilon_\theta$. For example, Stable Diffusion XL~\cite{podell2023sdxl} requires 6,763 GMACs per step to generate a $1024\times1024$ image. This computational demand escalates more than quadratically with increasing resolution, making the latency for generating a single high-resolution image impractically high for real-world applications. Furthermore, given that $\mathbf x_{t-1}$ depends on $\mathbf x_t$, parallel computation of $\epsilon_t$ and $\epsilon_{t-1}$ is challenging. Hence, even with multiple idle GPUs, accelerating the generation of a single high-resolution image remains tricky. Recently, Shih \etal introduced ParaDiGMS~\cite{shih2023paradigms}, employing Picard iterations to parallelize the denoising steps in a data-parallel manner. However, ParaDiGMS wastes the computation on speculative guesses that fail quality thresholds. It also relies on a large total step count $T$ to exploit multi-GPU data parallelism, limiting its potential applications. Another conventional method is sharding the model on multiple devices and using tensor parallelism for inference. However, this method suffers from intolerable communication costs, making it impractical for real-world applications. Beyond these two schemes, are there alternative strategies for distributing workloads across multiple GPU devices so that single-image generation can also enjoy the free-lunch speedups from multiple devices?

\begin{figure*}[t]
    \centering
    \includegraphics[width=\linewidth]{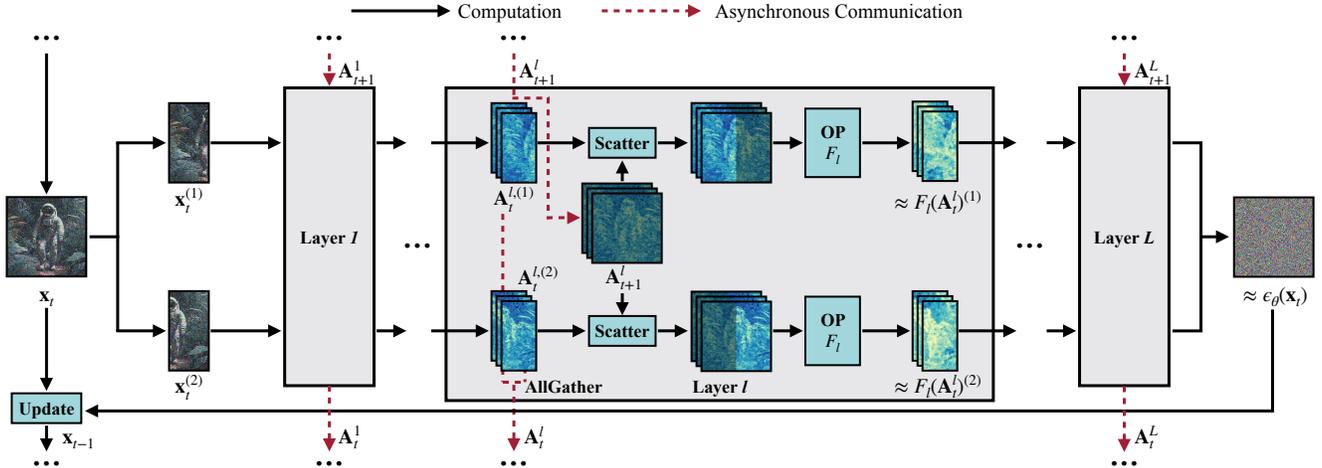}
    \vspace{-20pt}
	\caption{
        \looseness=-1
		Overview of \method. For simplicity, we omit the inputs of $t$ and $c$, and use $N=2$ devices as an example. Superscripts $^{(1)}$ and $^{(2)}$ represent the first and the second patch, respectively. Stale activations from the previous step are darkened. At each step $t$, we first split the input $\mathbf x_t$ into $N$ patches $\mathbf x_t^{(1)},\ldots,\mathbf x_t^{(N)}$. For each layer $l$ and device $i$, upon getting the input activation patches $\mathbf A_{t}^{l,(i)}$, two operations then process asynchronously: First, on device $i$, $\mathbf A_{t}^{l, (i)}$ is scattered back into the stale activation $\mathbf A_{t+1}^l$ from the previous step. The output of this \texttt{Scatter} operation is then fed into the sparse operator $F_l$ (linear, convolution, or attention layers), which performs computations exclusively on the fresh regions and produces the corresponding output. Meanwhile, an \texttt{AllGather} operation is performed over $\mathbf{A}_{t}^{l, (i)}$ to prepare the full activation $\mathbf{A}_{t}^l$ for the next step. We repeat this procedure for each layer. The final outputs are then aggregated together to approximate $\epsilon_\theta(\mathbf x_t)$, which is used to compute $\mathbf x_{t-1}$. The timeline visualization of each device for predicting $\epsilon_\theta(\mathbf x_t)$ is shown in \fig{timeline}. 
    }
    \vspace{-10pt}
    \lblfig{method}
\end{figure*}                                                                                                      
\begin{figure}[t]
    \centering
    \includegraphics[width=\linewidth]{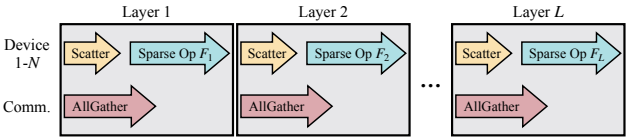}
    \vspace{-20pt}
	\caption{
        \looseness=-1
		Timeline visualization on each device when predicting $\epsilon_\theta(\mathbf x_t)$. \textit{Comm.} means communication, which is asynchronous with computation. The \texttt{AllGather} overhead is fully hidden within the computation. 
	}
    \vspace{-15pt}
    \lblfig{timeline}
\end{figure}
\section{Method}
\lblsect{Method}

\looseness=-1
The key idea of \method is to parallelize computation across devices by splitting the image into patches. \Naively, this can be done by either (1) independently computing patches and stitching them together, or (2) synchronously communicating intermediate activations between patches. However, the first approach leads to visible discrepancies at the boundaries of each patch due to the absence of interaction between them (see \fig{teaser} and \fig{idea}(b)). The second approach, on the other hand, incurs excessive communication overheads, negating the benefits of parallel processing. To address these challenges, we propose a novel parallelism paradigm, \emph{\parallelism}, which leverages the sequential nature of diffusion models to overlap communication and computation. Our key insight is reusing slightly outdated, or `stale' activations from the previous diffusion step to facilitate interactions between patches, which we describe as \emph{activation displacement}. This is based on the observation that the inputs for consecutive denoising steps are relatively similar. Consequently, computing each patch's activation at a layer does not rely on other patches' fresh activations, allowing communication to be hidden within subsequent layers' computation. We will next provide a detailed breakdown of each aspect of our algorithm and system design.

\myparagraph{\Parallelism.} As shown in \fig{method}, when predicting $\epsilon_\theta(\mathbf x_t)$ (we omit the inputs of timestep $t$ and condition $c$ here for simplicity), we first split $\mathbf x_t$ into multiple patches $\mathbf x_t^{(1)}, \mathbf x_t^{(2)}, \ldots, \mathbf x_t^{(N)}$, where $N$ is the number of devices. For example, we use $N=2$ in \fig{method}. Each device has a replicate of the model $\epsilon_\theta$ and will process a single patch independently, in parallel.

\looseness=-1
For a given layer $l$, let's consider the input activation patch on the $i$-th device, denoted as $A_t^{l, (i)}$. This patch is first scattered into the stale activations from the previous step, $A_{t+1}^l$, at its corresponding spatial location (the method for obtaining $A_{t+1}^l$ will be discussed later). Here, $A_{t+1}^l$ is in full spatial shape. In the \texttt{Scatter} output, only the $\frac{1}{N}$ regions where $A_t^{l, (i)}$ is placed are fresh and require recomputation. We then selectively apply the layer operation $F_l$ (linear, convolution, or attention) to these fresh areas, thereby generating the output for the corresponding regions. This process is repeated for each layer. Finally, the outputs from all layers are synchronized together to approximate $\epsilon_\theta(\mathbf x_t)$. Through this methodology, each device is responsible for only $\frac{1}{N}$ of the total computations, enabling efficient parallelization.

There still remains a problem of how to obtain the stale activations from the previous step. As shown in \fig{method}, at each timestep $t$, when device $i$ acquires $A_{t}^{l,(i)}$, it will then broadcast the activations to all other devices and perform the \texttt{AllGather} operation. Modern GPUs often support asynchronous communication and computation, which means that this \texttt{AllGather} process does not block ongoing computations. By the time we reach layer $l$ in the next timestep, each device should have already received a replicate of $A_{t}^l$. Such an approach effectively hides communication overheads within the computation phase, as shown in \fig{timeline}. However, there is an exception: the very first step (\ie, $\mathbf x_T$). In this scenario, each device simply executes the standard synchronous communication and caches the intermediate activations for the next step.

\myparagraph{Sparse operations.}
For each layer $l$, we modify the original operator $F_l$ to enable sparse computation selectively on the fresh areas. Specifically, if $F_l$ is a convolution, linear, or cross-attention layer, we apply the operator exclusively to the newly refreshed regions, rather than the full feature map. This can be achieved by extracting the fresh sections from the \texttt{scatter} output and feeding them into $F_l$. For layers where $F_l$ is a self-attention layer, we transform it into a cross-attention layer, similar to SIGE~\cite{li2022efficient}. In this setting, only the query tokens from the fresh areas are preserved on the device, while the key and value tokens still encompass the entire feature map (the \texttt{scatter} output). Thus, the computational cost for $F_l$ is exactly proportional to the size of the fresh area.

\looseness=-1
\myparagraph{Corrected asynchronous GroupNorm.}
Diffusion models often adopt group normalization (GN)~\cite{wu2018group,nichol2021improved} layers in the network. These layers normalize across the spatial dimension, necessitating the aggregation of activations to restore their full spatial shape. In \sect{Ablation Study}, we discover that either normalizing only the fresh patches or reusing stale features degrades image quality. However, aggregating all the normalization statistics will incur considerable overhead due to the synchronous communication. To solve this dilemma, we additionally introduce a correction term to the stale statistics. Specifically, for each device $i$ at a given step $t$, every GN layer can compute the group-wise mean of its fresh patch $\mathbf A_t^{(i)}$, denoted as $\mathbb E[\mathbf A_t^{(i)}]$. For simplicity, we omit the layer index $l$ here. It also has cached the local mean $\mathbb E[\mathbf A_{t+1}^{(i)}]$ and aggregated global mean $\mathbb E[\mathbf A_{t+1}]$ from the previous step. Then the approximated global mean $\mathbb E[\mathbf A_t]$ for current step on device $i$ can be computed as
\begin{equation}
	\mathbb E[\mathbf A_t] \approx \underbrace{\mathbb E[\mathbf A_{t+1}]}_{\text{stale global mean}} + \underbrace{(\mathbb E[\mathbf A_t^{(i)}] - \mathbb E[\mathbf A_{t+1}^{(i)}])}_{\text{correction}}.
\end{equation}
We use the same technique to approximate $\mathbb E[(\mathbf A_t)^2]$, then the variance can be approximated as $\mathbb E[(\mathbf A_t)^2]-\mathbb E[\mathbf A_t]^2$. We then use these approximated statistics for the GN layer and in the meantime aggregate the local mean and variance to compute the precise ones using asynchronous communication. Thus, the communication cost can also be pipelined into the computation. We empirically find this method yields comparable results to the direct synchronous aggregation. However, there are some rare cases where the approximated variance is negative. For these negative variance groups, we will fall back to use the local variance of the fresh patch.

\looseness=-1
\myparagraph{Warm-up steps.} As observed in eDiff-I~\cite{balaji2022ediffi} and FastComposer~\cite{xiao2023fastcomposer}, the behavior of diffusion synthesis undergoes qualitative changes throughout the denoising process. Specifically, the initial steps of sampling predominantly shape the low-frequency aspects of the image, such as spatial layout and overall semantics. As the sampling progresses, the focus shifts to recovering local high-frequency details. Therefore, to boost image quality, especially in samplers with a reduced number of steps, we adopt warm-up steps. Instead of directly employing \parallelism after the first step, we continue with several iterations of the standard synchronous patch parallelism as a preliminary phase, or \textit{warm-up}. As detailed in \sect{Few-step sampling and warm-up steps.}, this integration of warm-up steps significantly improves performance. 

\begin{figure*}[t]
    \centering
    \includegraphics[width=\linewidth]{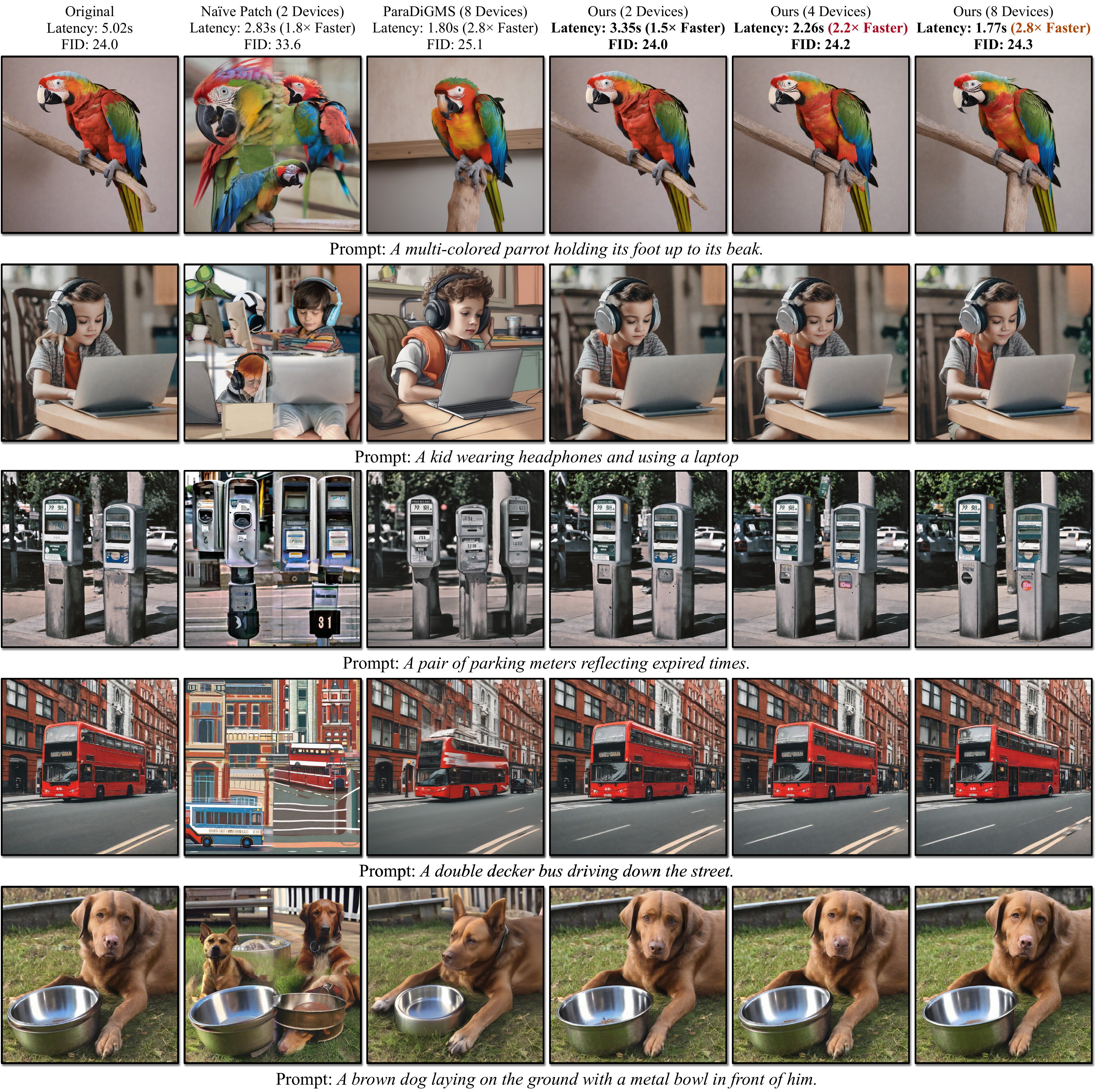}
    \vspace{-20pt}
	\caption{
		Qualitative results. FID is computed against the ground-truth images. Our \method can reduce the latency according to the number of used devices while preserving visual fidelity.
    }
    \vspace{-15pt}
    \lblfig{quality}
\end{figure*}

\section{Experiments}
We first describe our experiment setups, including our benchmark datasets, baselines, and evaluation protocols. Then we present our main results regarding both quality and efficiency. Finally, we further show some ablation studies to verify each design choice.
\subsection{Setups}

\looseness=-1 
\myparagraph{Models.} As our method only requires off-the-shelf pre-trained diffusion models, we mainly conduct experiments on the state-of-the-art public text-to-image model Stable Diffusion XL (SDXL)~\cite{podell2023sdxl}. SDXL first compresses an image to an $8\times$ smaller latent representation using a pre-trained auto-encoder and then applies a diffusion model in this latent space. It also incorporates multiple cross-attention layers to facilitate text conditioning. Compared to the original Stable Diffusion~\cite{rombach2022high}, SDXL adopts significantly more attention layers, resulting in a more computationally intensive model.

\looseness=-1
\myparagraph{Datasets.} We use the \href{https://huggingface.co/datasets/HuggingFaceM4/COCO}{HuggingFace version} of COCO Captions 2014~\cite{chen2015microsoft} dataset to benchmark our method. This dataset contains human-generated captions for images from Microsoft Common Objects in COntext (COCO) dataset~\cite{lin2014microsoft}. For evaluation, we randomly sample a subset from the validation set, which contains 5K images with one caption per image.

\renewcommand \arraystretch{1.}
\begin{table*}[t]
    \setlength{\tabcolsep}{7pt}
    \scriptsize \centering
    \begin{tabular}{cclcccccccc}
        \toprule
        \multirow{2}{*}[\multirowcenter]{\#Steps} & \multirow{2}{*}[\multirowcenter]{\#Devices} & \multirow{2}{*}[\multirowcenter]{Method} & \multirow{2}{*}[\multirowcenter]{PSNR ($\uparrow$)} & \multicolumn{2}{c}{LPIPS ($\downarrow$)}  & \multicolumn{2}{c}{FID ($\downarrow$)} & \multirow{2}{*}[\multirowcenter]{MACs (T)} & \multicolumn{2}{c}{Latency} \\
        \cmidrule(lr){5-6} \cmidrule(lr){7-8} \cmidrule(lr){10-11}
        & & & & w/ G.T. & w/ Orig. & w/ G.T. & w/ Orig. & & Value (s) & Speedup\\
        \midrule
        & 1 & Original & -- & 0.797 & -- & 24.0 & -- & 338 & 5.02 & --  \\
        \cmidrule[0.55pt]{2-11}
        & \multirow{2}{*}[\multirowcenter]{2} & \Naive Patch & 14.0 & 0.812 & 0.596 & 33.6 & 29.4 & \textbf{322} & \textbf{2.83} & \textbf{1.8}$\times$ \\
        \cmidrule{3-11}
        & & \textbf{Ours} & \textbf{24.6} & \textbf{0.797} & \textbf{0.146} & \textbf{24.2} & \textbf{4.86} & 338 & 3.35 & 1.5$\times$ \\
        \cmidrule[0.55pt]{2-11}
        \multirow{2}{*}[\multirowcenter]{50} & \multirow{2}{*}[\multirowcenter]{4} & \Naive Patch & 10.7 & 0.853 & 0.753 & 125 & 133 & \textbf{318} & \textbf{1.74} & \textbf{2.9}$\times$\\
        \cmidrule{3-11}
        & & \textbf{Ours} & \textbf{23.0} & \textbf{0.798} & \textbf{0.183} & \textbf{24.2} & \textbf{5.76} & 338 & 2.26 & 2.2$\times$ \\
        \cmidrule[0.55pt]{2-11}
        & & \Naive Patch & 7.70 & 0.892 & 0.857 & 252 & 259 & \textbf{324} & \textbf{1.27} & \textbf{4.0}$\times$ \\
        \cmidrule{3-11}
        & 8 & ParaDiGMS & 19.7 & 0.800 & 0.320 & 25.1 & 10.8 & 657 & 1.80 & 2.8$\times$ \\
        \cmidrule{3-11}
        & & \textbf{Ours} & \textbf{22.0} & \textbf{0.799} & \textbf{0.211} & \textbf{24.4} & \textbf{6.46} & 338 & 1.77 & 2.8$\times$ \\
        \midrule 
        & 1 & Original & -- & 0.801 & -- & 23.9 & -- & 169 & 2.52 & -- \\
        \cmidrule[0.55pt]{2-11}
        25 & \multirow{2}{*}[\multirowcenter]{8} & ParaDiGMS & 21.3 & 0.808 & 0.273 & 25.8 & 10.4 & 721 & 1.89 & 1.3$\times$ \\
        \cmidrule{3-11}
        & & \textbf{Ours} & \textbf{24.7} & \textbf{0.802} & \textbf{0.161} & \textbf{24.6} & \textbf{5.67} & \textbf{169} & \textbf{0.93} & \textbf{2.7}$\times$ \\
        \bottomrule
    \end{tabular}
    \vspace{-5pt}
    \caption{
    	Quantitative evaluation. \textit{MACs} measures cumulative computation across all devices for the whole denoising process for generating a single $1024\times1024$ image. \textit{w/ G.T.} means calculating the metrics with the ground-truth images, while \textit{w/ Orig.} means with the original model's samples. For PSNR, we report the \textit{w/ Orig.} setting. Our method mirrors the results of the original model across all metrics while maintaining the total MACs. It also reduces the latency on NVIDIA A100 GPUs in proportion to the number of used devices.
    }
    \lbltab{quantitative}
    \vspace{-10pt}
\end{table*}

\myparagraph{Baselines.} We compare our \method against the following baselines in terms of both quality and efficiency:
\begin{itemize}[leftmargin=*]
    \item \textit{\Naive Patch}. At each iteration, the input is divided row-wise or column-wise alternately. These patches are then processed independently by the model, without any interaction between them. The outputs are subsequently concatenated together.
    \item \textit{ParaDiGMS~\cite{shih2023paradigms}} is a technique to accelerate pre-trained diffusion models by denoising multiple steps in parallel. It uses Picard iterations to guess the solution of future steps and iteratively refines it until convergence. We use a batch size 8 for ParaDiGMS to align with Table 4 in the original paper~\cite{shih2023paradigms}. We empirically find this setting yields the best performance in both quality and latency.
\end{itemize}

\myparagraph{Metrics.} 
Following previous works~\cite{meng2022sdedit,li2020gan,park2019semantic,li2022efficient}, we evaluate the image quality with standard metrics: Peak Signal Noise Ratio (PSNR, higher is better), LPIPS (lower is better)~\cite{zhang2018perceptual}, and \fid (FID, lower is better)~\cite{heusel2017gans}\footnote{We use \href{https://torchmetrics.readthedocs.io/en/stable/}{TorchMetrics} to calculate PSNR and LPIPS, and use CleanFID~\citep{DBLP:conf/cvpr/Parmar0Z22} to calculate FID.}. We employ PSNR to quantify the minor numerical differences between the outputs of the benchmarked method and the original diffusion model outputs. LPIPS is used to evaluate perceptual similarity. Additionally, the FID score is used to measure the distributional differences between the outputs of the method and either the original outputs or the ground-truth images.

\myparagraph{Implementation details.} 
By default, we adopt the 50-step DDIM sampler~\cite{song2020denoising} with classifier-free guidance scale 5 to generate $1024\times1024$ images, unless otherwise specified. In addition to the first step, we perform another 4-step synchronous patch parallelism, serving as a warm-up phase.

\looseness=-1
We use PyTorch 2.2~\cite{paszke2019pytorch} to benchmark the speedups of our method. To measure latency, we first warm up the devices with 3 iterations of the whole denoising process, then run another 10 iterations and calculate the average latency by discarding the results of the fastest and slowest runs. Additionally, we use CUDAGraph to optimize some kernel launching overhead for both the original model and our method.

\begin{figure}[t]
    \centering
    \includegraphics[width=\linewidth]{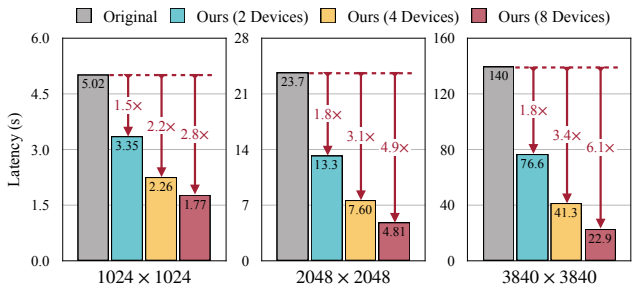}
    \vspace{-20pt}
	\caption{
        Measured total latency of \method with the 50-step DDIM sampler~\cite{song2020denoising} for generating a single image across different resolutions on NVIDIA A100 GPUs. When scaling up the resolution, the GPU devices are better utilized. Remarkably, when generating $3840\times3840$ images, \method achieves 1.8$\times$, 3.4$\times$ and 6.1$\times$ speedups with 2, 4, and 8 A100s, respectively.
	}
    \vspace{-10pt}
    \lblfig{speedups}
\end{figure}

\subsection{Main Results}
\lblsect{Main Results}

\myparagraph{Quality results.}
In \fig{quality}, we show some qualitative visual results and report some quantitative evaluation in \tab{quantitative}. \textit{with G.T.} means computing the metric with the ground-truth COCO~\cite{lin2014microsoft} images, whereas \textit{w/ Orig.} refers to computing the metrics with the outputs from the original model. For PSNR, we report only the \textit{w/ Orig.} setting, as the \textit{w/ G.T.} comparison is not informative due to significant numerical differences between the generated outputs and the ground-truth images.

\looseness=-1
As shown in \tab{quantitative}, ParaDiGMS~\cite{shih2023paradigms} expends considerable computational resources on guessing future denoising steps, resulting in a much higher total MACs. Besides, it also suffers from some performance degradation. In contrast, our method simply distributes workloads across multiple GPUs, maintaining a constant total computation. The \Naive Patch baseline, while lower in total MACs, lacks the crucial inter-patch interaction, leading to fragmented outputs. This limitation significantly impacts image quality, as reflected across all evaluation metrics. Our \method can well preserve interaction. Even when using 8 devices, it achieves comparable PSNR, LPIPS, and FID scores comparable to those of the original model.

\myparagraph{Speedups.} 
Compared to the theoretical computation reduction, on-hardware acceleration is more critical for real-world applications. To demonstrate the effectiveness of our method, we also report the end-to-end latency in \tab{quantitative} on 8 NVIDIA A100 GPUs. In the 50-step setting, ParaDiGMS achieves an identical speedup of $2.8\times$ to our method at the cost of compromised image quality (see \fig{quality}). In the more commonly used 25-step setting, ParaDiGMS only has a marginal $1.3\times$ speedup due to excessive wasted guesses, which is also reported in Shih \etal~\cite{shih2023paradigms}. However, our method can still mirror the original quality and accelerate the model by 2.7$\times$.

\looseness=-1 When generating $1024\times1024$ images, our speedups are limited by the low GPU utilization of SDXL. To maximize device usage, we further scale the resolution to $2048\times2048$ and $3840\times3840$ in \fig{speedups}. At these larger resolutions, the GPU devices are better utilized. Specifically, for $3840\times3840$ images, \method reduces the latency by 1.8$\times$, 3.4$\times$ and 6.1$\times$ with 2, 4 and 8 A100s, respectively. Note that these results are benchmarked with PyTorch. With more advanced compilers, such as TVM~\cite{chen2018tvm} and TensorRT~\cite{Rao2023NVIDIA}, we anticipate even higher GPU utilization and consequently more pronounced speedups from \method, as observed in SIGE~\cite{li2022efficient}. In practical use, the batch size often doubles due to classifier-free guidance~\cite{ho2021classifier}. We can first split the batch and then apply \method to each batch separately. This approach further improves the total speedups to 3.6$\times$ and 6.6$\times$ with 4 and 8 A100s for generating a single $3840\times3840$ image, respectively.

\subsection{Ablation Study}
\lblsect{Ablation Study}
\lblsect{Communication cost.}
\myparagraph{Compare to tensor parallelism.} 
In \tab{communication}, we benchmark our latency with synchronous tensor parallelism (\textit{Sync. TP}) and synchronous patch parallelism (\textit{Sync. PP}), and report the corresponding communication amounts. Compared to TP, PP has better independence, which eliminates the need for communication within cross-attention and linear layers. For convolutional layers, communication is only required at the patch boundaries, which represent a minimal portion of the entire tensor. Moreover, PP utilizes \texttt{AllGather} over \texttt{AllReduce}, leading to lower communication demands and no additional use of computing resources. Therefore, PP requires $60\%$ fewer communication amounts and is $1.6\sim2.1\times$ faster than TP, making it a more efficient approach for deploying diffusion models. We also include a theoretical PP baseline without any communication (\textit{No Comm.}) to demonstrate the communication overhead in \textit{Sync. PP} and \method. Compared to \textit{Sync. PP}, \method further cuts such overhead by over $50\%$. The remaining overhead mainly comes from our current usage of NVIDIA Collective Communication Library (NCCL) for asynchronous communication. NCCL kernels use SMs (the computing resources on GPUs), which will slow down the overlapped computation. Using remote memory access can bypass this issue and close the performance gap.

 \renewcommand \arraystretch{1.}
 \begin{table}[t]
     \setlength{\tabcolsep}{3.2pt}
     \scriptsize \centering
     \begin{tabular}{lcccccc}
 		\toprule
 		\multirow{2}{*}[\multirowcenter]{Method} & \multicolumn{2}{c}{$1024\times1024$} & \multicolumn{2}{c}{$2048\times2048$} & \multicolumn{2}{c}{$3840\times3840$} \\
 		\cmidrule(lr){2-3}\cmidrule(lr){4-5}\cmidrule(lr){6-7}
 		& Comm. & Latency & Comm. & Latency & Comm. & Latency \\
 		\midrule
 		Original & -- & 5.02s & -- & 23.7s & -- & 140s \\
 		\midrule 
 		Sync. TP & 1.33G & 3.61s & 5.33G & 11.7s & 18.7G & 46.3s \\
 		Sync. PP & 0.42G & 2.21s & 1.48G & 5.62s & 5.38G & 24.7s \\
 		\textbf{\method (Ours)} & \textbf{0.42G} & \textbf{1.77s} & \textbf{1.48G} & \textbf{4.81s} & \textbf{5.38G} & \textbf{22.9s} \\
        \midrule
 		No Comm. & -- & 1.48s & -- & 4.14s & -- & 21.3s \\
		\bottomrule
     \end{tabular}
     \vspace{-5pt}
     \caption{
        \looseness=-1
        Communication cost comparisons with 8 A100s across different resolutions. \textit{Sync. TP/PP}: Synchronous tensor/patch parallelism. \textit{No Comm.}: An ideal no communication PP. \textit{Comm.} measures the total communication amount. PP only requires less than $\frac{1}{3}$ communication amounts compared to TP. Our \method further reduces the communication overhead by $50\sim60\%$.
    }
    \lbltab{communication}
    \vspace{-15pt}
 \end{table}

\begin{figure}[t]
    \centering
    \vspace{-5pt}
    \includegraphics[width=\linewidth]{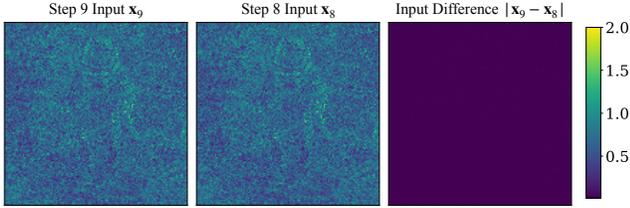}
    \vspace{-18pt}
	\caption{
		Visualization of the inputs from steps 9 and 8 and their difference. All feature maps are channel-wise averaged. The difference is nearly all zero, exhibiting high similarity.
	}
    \vspace{-10pt}
    \lblfig{similarity}
\end{figure}
\myparagraph{Input similarity.} Our \parallelism relies on the assumption that the inputs from consecutive denoising steps are similar. To support this claim, we quantitatively calculate the model input difference across all consecutive steps using a 50-step DDIM sampler. The average difference is only 0.02, within the input range of $[-4,4]$ (about $0.3\%$). \fig{similarity} further qualitatively visualizes the input difference between steps 9 and 8 (randomly selected). The difference is nearly all zero, substantiating our hypothesis of high similarity between inputs from neighboring steps.

\lblsect{Few-step sampling and warm-up steps.}
\myparagraph{Few-step sampling and warm-up steps.} As stated above, our approach hinges on the observation that adjacent denoising steps share similar inputs, \ie, $\mathbf x_t \approx \mathbf x_{t-1}$. However, as we increase the step size and thereby reduce the number of steps, the approximation error escalates, potentially compromising the effectiveness of our method. In \fig{few step sampling}, we present results using 10-step DPM-Solver~\cite{lu2022dpm,lu2022dpm++}. The 10-step configuration is the threshold for the training-free samplers to maintain the image quality. Under this setting, \naive \method without warm-up struggles to preserve the image quality. However, incorporating an additional two-step warm-up significantly recovers the performance with only slightly increased latency. 

\begin{figure}[t]
    \centering
    \includegraphics[width=\linewidth]{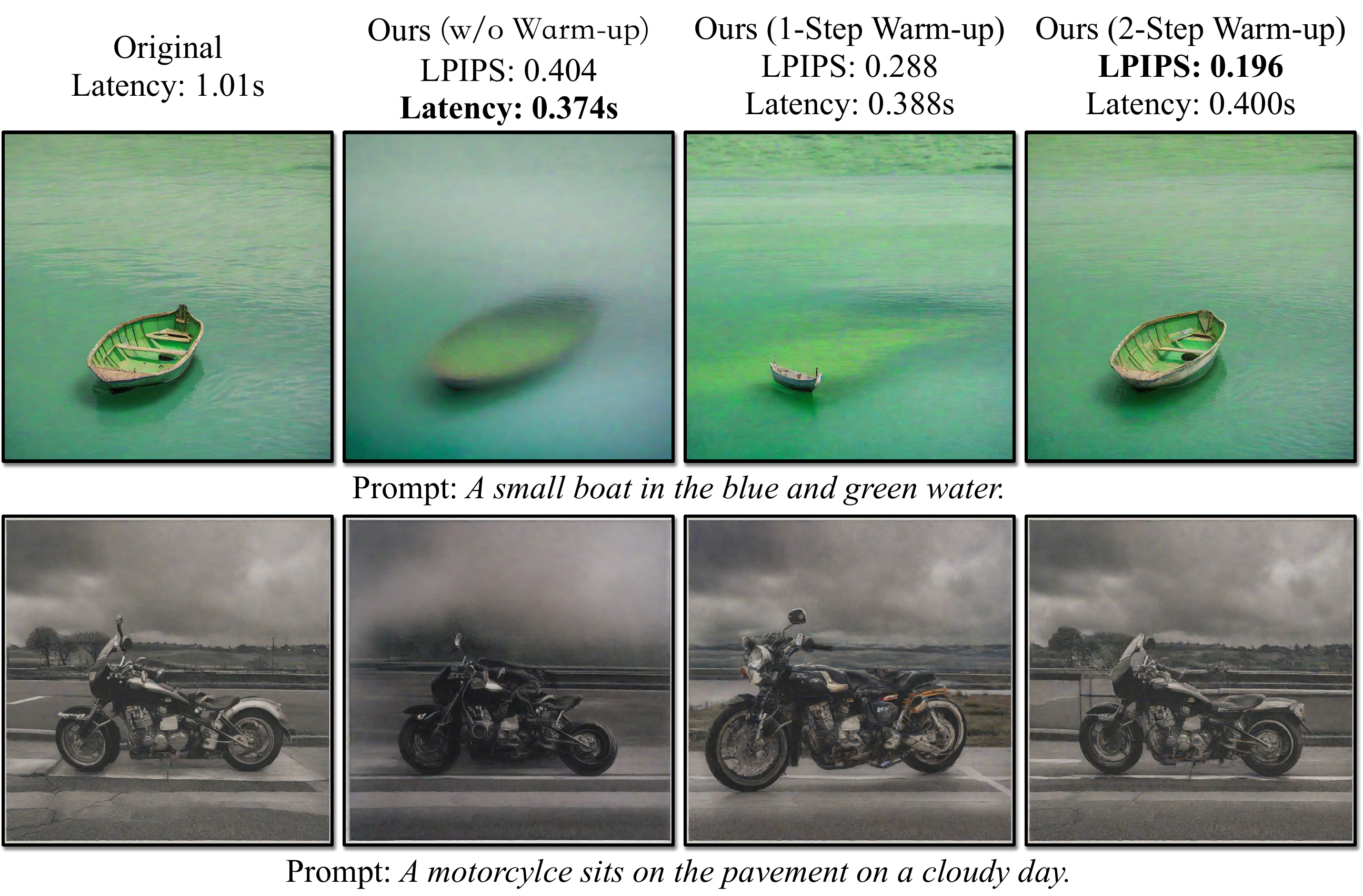}
    \vspace{-20pt}
	\caption{
        \looseness=-1 Qualitative results on the 10-step DPM-Solver~\cite{lu2022dpm,lu2022dpm++} with different warm-up steps. LPIPS is computed against the samples from the original SDXL over the entire COCO~\cite{chen2015microsoft} dataset. \Naive \method without warm-up steps has evident quality degradation. Adding a 2-step warm-up significantly improves the performance while avoiding high latency rise.
	}
    \vspace{-15pt}
    \lblfig{few step sampling}
\end{figure}

\looseness=-1
\myparagraph{GroupNorm.} As discussed in \sect{Method}, calculating accurate group normalization (GN) statistics is crucial for preserving image quality. In \fig{gn}, we compare four different GN schemes. The first approach \textit{Separate GN} uses statistics from the on-device fresh patch. This approach delivers the best speed at the cost of lower image fidelity. This compromise is particularly severe for large numbers of used devices, due to insufficient patch size for precise statistics estimation. The second scheme \textit{Stale GN} computes statistics using stale activations. However, this method also faces quality degradation, because of the different distributions between stale and fresh activations, often resulting in images with a fog-like noise effect. The third approach \textit{Sync. GN} use synchronized communication to aggregate accurate statistics. Though achieving the best image quality, it suffers from large synchronization overhead. Our method uses a correction term to close the distribution gap between the stale and fresh statistics. It achieves image quality on par with \textit{Sync. GN} but without incurring synchronous communication overhead.

\begin{figure}[t]
    \centering
    \vspace{-5pt}
    \includegraphics[width=\linewidth]{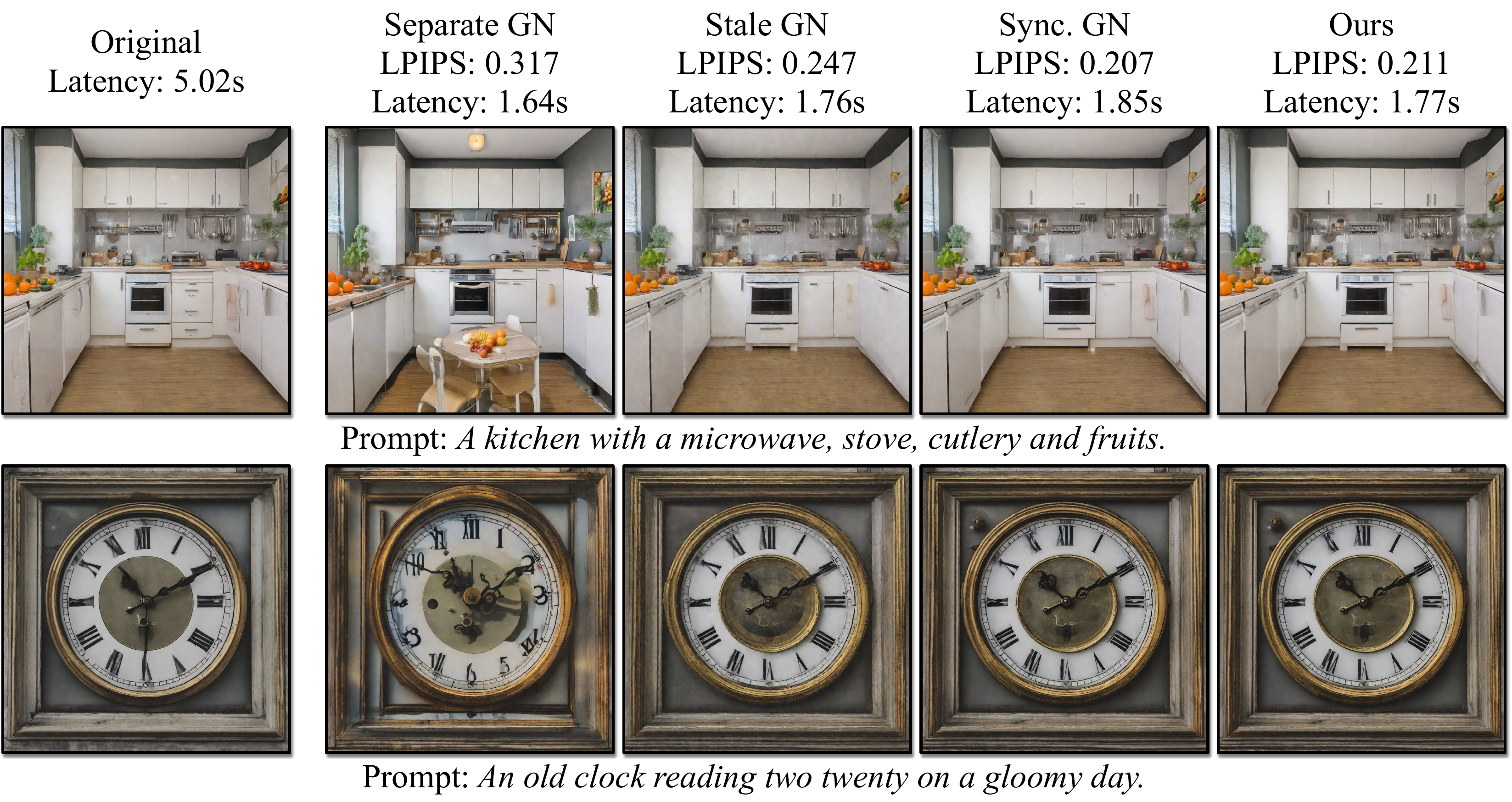}
    \vspace{-20pt}
	\caption{
		Qualitative results of different GN schemes with 8 A100s. LPIPS is computed against the original samples over the whole COCO~\cite{chen2015microsoft} dataset. \textit{Separate GN} only utilizes the statistics from the on-device patch. \textit{Stale GN} reuses the stale statistics. They suffer from quality degradation. \textit{Sync. GN} synchronizes data to ensure accurate statistics at the cost of extra overhead. Our corrected asynchronous GN, by correcting stale statistics, avoids the need for synchronization and effectively restores quality.
	}
    \vspace{-15pt}
    \lblfig{gn}
\end{figure}

\section{Conclusion \& Discussion}
In this paper, we introduce \method to accelerate diffusion models with multiple GPUs for parallelism. Our method divides images into patches, assigning each to a separate GPU. We reuse the pre-computed activations from previous steps to maintain patch interactions. On Stable Diffusion XL, our method achieves up to a 6.1$\times$ speedup on 8 NVIDIA A100s. This advancement not only enhances the efficiency of AI-generated content creation but also sets a new benchmark for future research in parallel computing for AI applications.

\myparagraph{Limitations.} 
To fully hide the communication overhead within the computation, NVLink is essential for \method to maximize the speedup. However, NVLink has been widely used recently. Moreover, quantization~\cite{li2023q} can also reduce the communication workloads for our method. Besides, \method has limited speedups for low-resolution images as the devices are underutilized. Advanced compilers~\cite{Rao2023NVIDIA,chen2018tvm} would help to exploit the devices and achieve better speedups. Our method may not work for the extremely-few-step methods~\cite{salimans2021progressive,song2023consistency,meng2022distillation,luo2023latent,luo2023lcmlora}, due to the rapid changes of the denoising states. Yet our preliminary experiment suggests that slightly more steps (\eg, 10) are enough for \method to obtain high-quality results.

%
\subsubsection*{Acknowledgments}
We thank Jun-Yan Zhu and Ligeng Zhu for their helpful discussion and valuable feedback. The project is supported by MIT-IBM Watson AI Lab, Amazon, MIT Science Hub, and National Science Foundation.

\subsubsection*{Changelog}
\noindent\textbf{V1} Initial preprint release (CVPR 2024).

\noindent\textbf{V2} Update Figures~\ref{fig:teaser} and \ref{fig:idea}.

\noindent\textbf{V3} Correct the PSNR values in \tab{quantitative}.

{
    \small
    \bibliographystyle{ieeenat_fullname}
    \bibliography{main}

\begin{thebibliography}{78}
\providecommand{\natexlab}[1]{#1}
\providecommand{\url}[1]{\texttt{#1}}
\expandafter\ifx\csname urlstyle\endcsname\relax
  \providecommand{\doi}[1]{doi: #1}\else
  \providecommand{\doi}{doi: \begingroup \urlstyle{rm}\Url}\fi

\bibitem[Rao(2023)]{Rao2023NVIDIA}
\emph{NVIDIA/TensorRT}.
\newblock 2023.

\bibitem[Balaji et~al.(2022)Balaji, Nah, Huang, Vahdat, Song, Kreis, Aittala, Aila, Laine, Catanzaro, et~al.]{balaji2022ediffi}
Yogesh Balaji, Seungjun Nah, Xun Huang, Arash Vahdat, Jiaming Song, Karsten Kreis, Miika Aittala, Timo Aila, Samuli Laine, Bryan Catanzaro, et~al.
\newblock ediffi: Text-to-image diffusion models with an ensemble of expert denoisers.
\newblock \emph{arXiv preprint arXiv:2211.01324}, 2022.

\bibitem[Betker et~al.(2023)Betker, Goh, Jing, Brooks, Wang, Li, Ouyang, Zhuang, Lee, Guo, et~al.]{betker2023improving}
James Betker, Gabriel Goh, Li Jing, Tim Brooks, Jianfeng Wang, Linjie Li, Long Ouyang, Juntang Zhuang, Joyce Lee, Yufei Guo, et~al.
\newblock Improving image generation with better captions.
\newblock \emph{Computer Science. https://cdn.openai.com/papers/dall-e-3.pdf}, 2023.

\bibitem[Chen et~al.(2018)Chen, Moreau, Jiang, Zheng, Yan, Shen, Cowan, Wang, Hu, Ceze, et~al.]{chen2018tvm}
Tianqi Chen, Thierry Moreau, Ziheng Jiang, Lianmin Zheng, Eddie Yan, Haichen Shen, Meghan Cowan, Leyuan Wang, Yuwei Hu, Luis Ceze, et~al.
\newblock $\{$TVM$\}$: An automated $\{$End-to-End$\}$ optimizing compiler for deep learning.
\newblock In \emph{OSDI}, 2018.

\bibitem[Chen et~al.(2015)Chen, Fang, Lin, Vedantam, Gupta, Doll{\'a}r, and Zitnick]{chen2015microsoft}
Xinlei Chen, Hao Fang, Tsung-Yi Lin, Ramakrishna Vedantam, Saurabh Gupta, Piotr Doll{\'a}r, and C~Lawrence Zitnick.
\newblock Microsoft coco captions: Data collection and evaluation server.
\newblock \emph{arXiv preprint arXiv:1504.00325}, 2015.

\bibitem[Dai et~al.(2023)Dai, Hou, Ma, Tsai, Wang, Wang, Zhang, Vandenhende, Wang, Dubey, et~al.]{dai2023emu}
Xiaoliang Dai, Ji Hou, Chih-Yao Ma, Sam Tsai, Jialiang Wang, Rui Wang, Peizhao Zhang, Simon Vandenhende, Xiaofang Wang, Abhimanyu Dubey, et~al.
\newblock Emu: Enhancing image generation models using photogenic needles in a haystack.
\newblock \emph{arXiv preprint arXiv:2309.15807}, 2023.

\bibitem[Dong et~al.(2017)Dong, Huang, Yang, and Yan]{dong2017more}
Xuanyi Dong, Junshi Huang, Yi Yang, and Shuicheng Yan.
\newblock More is less: A more complicated network with less inference complexity.
\newblock In \emph{CVPR}, 2017.

\bibitem[Goodfellow et~al.(2014)Goodfellow, Pouget-Abadie, Mirza, Xu, Warde-Farley, Ozair, Courville, and Bengio]{goodfellow2014generative}
Ian Goodfellow, Jean Pouget-Abadie, Mehdi Mirza, Bing Xu, David Warde-Farley, Sherjil Ozair, Aaron Courville, and Yoshua Bengio.
\newblock Generative adversarial nets.
\newblock \emph{NeurIPS}, 2014.

\bibitem[Gu et~al.(2023)Gu, Zhai, Zhang, Susskind, and Jaitly]{gu2023matryoshka}
Jiatao Gu, Shuangfei Zhai, Yizhe Zhang, Josh Susskind, and Navdeep Jaitly.
\newblock Matryoshka diffusion models.
\newblock \emph{arXiv preprint arXiv:2310.15111}, 2023.

\bibitem[Han et~al.(2015)Han, Pool, Tran, and Dally]{han2015learning}
Song Han, Jeff Pool, John Tran, and William Dally.
\newblock Learning both weights and connections for efficient neural network.
\newblock \emph{NeurIPS}, 2015.

\bibitem[Heusel et~al.(2017)Heusel, Ramsauer, Unterthiner, Nessler, and Hochreiter]{heusel2017gans}
Martin Heusel, Hubert Ramsauer, Thomas Unterthiner, Bernhard Nessler, and Sepp Hochreiter.
\newblock Gans trained by a two time-scale update rule converge to a local nash equilibrium.
\newblock \emph{NeurIPS}, 2017.

\bibitem[Ho and Salimans(2021)]{ho2021classifier}
Jonathan Ho and Tim Salimans.
\newblock Classifier-free diffusion guidance.
\newblock In \emph{NeurIPS 2021 Workshop on Deep Generative Models and Downstream Applications}, 2021.

\bibitem[Ho et~al.(2020)Ho, Jain, and Abbeel]{ho2020denoising}
Jonathan Ho, Ajay Jain, and Pieter Abbeel.
\newblock Denoising diffusion probabilistic models.
\newblock \emph{NeurIPS}, 2020.

\bibitem[Hoogeboom et~al.(2023)Hoogeboom, Heek, and Salimans]{hoogeboom2023simple}
Emiel Hoogeboom, Jonathan Heek, and Tim Salimans.
\newblock simple diffusion: End-to-end diffusion for high resolution images.
\newblock \emph{arXiv preprint arXiv:2301.11093}, 2023.

\bibitem[Huang et~al.(2019)Huang, Cheng, Bapna, Firat, Chen, Chen, Lee, Ngiam, Le, Wu, et~al.]{huang2019gpipe}
Yanping Huang, Youlong Cheng, Ankur Bapna, Orhan Firat, Dehao Chen, Mia Chen, HyoukJoong Lee, Jiquan Ngiam, Quoc~V Le, Yonghui Wu, et~al.
\newblock Gpipe: Efficient training of giant neural networks using pipeline parallelism.
\newblock \emph{NeurIPS}, 2019.

\bibitem[Jaderberg et~al.(2014)Jaderberg, Vedaldi, and Zisserman]{jaderberg2014speeding}
Max Jaderberg, Andrea Vedaldi, and Andrew Zisserman.
\newblock Speeding up convolutional neural networks with low rank expansions.
\newblock In \emph{BMVC}, 2014.

\bibitem[Jia et~al.(2019)Jia, Zaharia, and Aiken]{jia2019beyond}
Zhihao Jia, Matei Zaharia, and Alex Aiken.
\newblock Beyond data and model parallelism for deep neural networks.
\newblock \emph{MLSys}, 2019.

\bibitem[Judd et~al.(2017)Judd, Delmas, Sharify, and Moshovos]{judd2017cnvlutin2}
Patrick Judd, Alberto Delmas, Sayeh Sharify, and Andreas Moshovos.
\newblock Cnvlutin2: Ineffectual-activation-and-weight-free deep neural network computing.
\newblock \emph{arXiv preprint arXiv:1705.00125}, 2017.

\bibitem[Kim and Ye(2021)]{kim2021diffusionclip}
Gwanghyun Kim and Jong~Chul Ye.
\newblock Diffusionclip: Text-guided image manipulation using diffusion models.
\newblock \emph{arXiv preprint arXiv:2110.02711}, 2021.

\bibitem[Kong and Ping(2021)]{kong2021fast}
Zhifeng Kong and Wei Ping.
\newblock On fast sampling of diffusion probabilistic models.
\newblock In \emph{ICML Workshop on Invertible Neural Networks, Normalizing Flows, and Explicit Likelihood Models}, 2021.

\bibitem[Li et~al.(2016)Li, Kadav, Durdanovic, Samet, and Graf]{li2016pruning}
Hao Li, Asim Kadav, Igor Durdanovic, Hanan Samet, and Hans~Peter Graf.
\newblock Pruning filters for efficient convnets.
\newblock \emph{ICLR}, 2016.

\bibitem[Li et~al.(2020)Li, Lin, Ding, Liu, Zhu, and Han]{li2020gan}
Muyang Li, Ji Lin, Yaoyao Ding, Zhijian Liu, Jun-Yan Zhu, and Song Han.
\newblock Gan compression: Efficient architectures for interactive conditional gans.
\newblock In \emph{CVPR}, 2020.

\bibitem[Li et~al.(2022)Li, Lin, Meng, Ermon, Han, and Zhu]{li2022efficient}
Muyang Li, Ji Lin, Chenlin Meng, Stefano Ermon, Song Han, and Jun-Yan Zhu.
\newblock Efficient spatially sparse inference for conditional gans and diffusion models.
\newblock In \emph{NeurIPS}, 2022.

\bibitem[Li et~al.(2017)Li, Liu, Luo, Change~Loy, and Tang]{li2017not}
Xiaoxiao Li, Ziwei Liu, Ping Luo, Chen Change~Loy, and Xiaoou Tang.
\newblock Not all pixels are equal: Difficulty-aware semantic segmentation via deep layer cascade.
\newblock In \emph{CVPR}, 2017.

\bibitem[Li et~al.(2023{\natexlab{a}})Li, Lian, Liu, Yang, Dong, Kang, Zhang, and Keutzer]{li2023q}
Xiuyu Li, Long Lian, Yijiang Liu, Huanrui Yang, Zhen Dong, Daniel Kang, Shanghang Zhang, and Kurt Keutzer.
\newblock Q-diffusion: Quantizing diffusion models.
\newblock \emph{arXiv preprint arXiv:2302.04304}, 2023{\natexlab{a}}.

\bibitem[Li et~al.(2023{\natexlab{b}})Li, Wang, Jin, Hu, Chemerys, Fu, Wang, Tulyakov, and Ren]{li2023snapfusion}
Yanyu Li, Huan Wang, Qing Jin, Ju Hu, Pavlo Chemerys, Yun Fu, Yanzhi Wang, Sergey Tulyakov, and Jian Ren.
\newblock Snapfusion: Text-to-image diffusion model on mobile devices within two seconds.
\newblock \emph{NeurIPS}, 2023{\natexlab{b}}.

\bibitem[Li et~al.(2021)Li, Zhuang, Guo, Zhuo, Zhang, Song, and Stoica]{li2021terapipe}
Zhuohan Li, Siyuan Zhuang, Shiyuan Guo, Danyang Zhuo, Hao Zhang, D. Song, and I. Stoica.
\newblock Terapipe: Token-level pipeline parallelism for training large-scale language models.
\newblock \emph{ICML}, 2021.

\bibitem[Li et~al.(2023{\natexlab{c}})Li, Zheng, Zhong, Liu, Sheng, Jin, Huang, Chen, Zhang, Gonzalez, and Stoica]{li2023alpaserve}
Zhuohan Li, Lianmin Zheng, Yinmin Zhong, Vincent Liu, Ying Sheng, Xin Jin, Yanping Huang, Z. Chen, Hao Zhang, Joseph~E. Gonzalez, and I. Stoica.
\newblock Alpaserve: Statistical multiplexing with model parallelism for deep learning serving.
\newblock \emph{USENIX Symposium on Operating Systems Design and Implementation}, 2023{\natexlab{c}}.

\bibitem[Lin et~al.(2021)Lin, Chen, Cai, Gan, and Han]{lin2021mcunetv2}
Ji Lin, Wei-Ming Chen, Han Cai, Chuang Gan, and Song Han.
\newblock Mcunetv2: Memory-efficient patch-based inference for tiny deep learning.
\newblock In \emph{Annual Conference on Neural Information Processing Systems (NeurIPS)}, 2021.

\bibitem[Lin et~al.(2014)Lin, Maire, Belongie, Hays, Perona, Ramanan, Doll{\'a}r, and Zitnick]{lin2014microsoft}
Tsung-Yi Lin, Michael Maire, Serge Belongie, James Hays, Pietro Perona, Deva Ramanan, Piotr Doll{\'a}r, and C~Lawrence Zitnick.
\newblock Microsoft coco: Common objects in context.
\newblock In \emph{Computer Vision--ECCV 2014: 13th European Conference, Zurich, Switzerland, September 6-12, 2014, Proceedings, Part V 13}, pages 740--755. Springer, 2014.

\bibitem[Liu et~al.(2015)Liu, Wang, Foroosh, Tappen, and Pensky]{liu2015sparse}
Baoyuan Liu, Min Wang, Hassan Foroosh, Marshall Tappen, and Marianna Pensky.
\newblock Sparse convolutional neural networks.
\newblock In \emph{CVPR}, 2015.

\bibitem[Lu et~al.(2022{\natexlab{a}})Lu, Zhou, Bao, Chen, Li, and Zhu]{lu2022dpm}
Cheng Lu, Yuhao Zhou, Fan Bao, Jianfei Chen, Chongxuan Li, and Jun Zhu.
\newblock Dpm-solver: A fast ode solver for diffusion probabilistic model sampling in around 10 steps.
\newblock \emph{arXiv preprint arXiv:2206.00927}, 2022{\natexlab{a}}.

\bibitem[Lu et~al.(2022{\natexlab{b}})Lu, Zhou, Bao, Chen, Li, and Zhu]{lu2022dpm++}
Cheng Lu, Yuhao Zhou, Fan Bao, Jianfei Chen, Chongxuan Li, and Jun Zhu.
\newblock Dpm-solver++: Fast solver for guided sampling of diffusion probabilistic models.
\newblock \emph{arXiv preprint arXiv:2211.01095}, 2022{\natexlab{b}}.

\bibitem[Luo et~al.(2023{\natexlab{a}})Luo, Tan, Huang, Li, and Zhao]{luo2023latent}
Simian Luo, Yiqin Tan, Longbo Huang, Jian Li, and Hang Zhao.
\newblock Latent consistency models: Synthesizing high-resolution images with few-step inference.
\newblock \emph{arXiv preprint arXiv: 2310.04378}, 2023{\natexlab{a}}.

\bibitem[Luo et~al.(2023{\natexlab{b}})Luo, Tan, Patil, Gu, von Platen, Passos, Huang, Li, and Zhao]{luo2023lcmlora}
Simian Luo, Yiqin Tan, Suraj Patil, Daniel Gu, Patrick von Platen, Apolinário Passos, Longbo Huang, Jian Li, and Hang Zhao.
\newblock Lcm-lora: A universal stable-diffusion acceleration module.
\newblock \emph{arXiv preprint arXiv: 2311.05556}, 2023{\natexlab{b}}.

\bibitem[Meng et~al.(2022{\natexlab{a}})Meng, Gao, Kingma, Ermon, Ho, and Salimans]{meng2022distillation}
Chenlin Meng, Ruiqi Gao, Diederik~P Kingma, Stefano Ermon, Jonathan Ho, and Tim Salimans.
\newblock On distillation of guided diffusion models.
\newblock \emph{arXiv preprint arXiv:2210.03142}, 2022{\natexlab{a}}.

\bibitem[Meng et~al.(2022{\natexlab{b}})Meng, He, Song, Song, Wu, Zhu, and Ermon]{meng2022sdedit}
Chenlin Meng, Yutong He, Yang Song, Jiaming Song, Jiajun Wu, Jun-Yan Zhu, and Stefano Ermon.
\newblock {SDE}dit: Guided image synthesis and editing with stochastic differential equations.
\newblock In \emph{ICLR}, 2022{\natexlab{b}}.

\bibitem[Narayanan et~al.(2019)Narayanan, Harlap, Phanishayee, Seshadri, Devanur, Ganger, Gibbons, and Zaharia]{narayanan2019pipedream}
Deepak Narayanan, Aaron Harlap, Amar Phanishayee, Vivek Seshadri, Nikhil~R Devanur, Gregory~R Ganger, Phillip~B Gibbons, and Matei Zaharia.
\newblock Pipedream: Generalized pipeline parallelism for dnn training.
\newblock In \emph{SOSP}, 2019.

\bibitem[Narayanan et~al.(2021)Narayanan, Shoeybi, Casper, LeGresley, Patwary, Korthikanti, Vainbrand, Kashinkunti, Bernauer, Catanzaro, Phanishayee, and Zaharia]{narayanan2021efficient}
D. Narayanan, M. Shoeybi, J. Casper, P. LeGresley, M. Patwary, V. Korthikanti, Dmitri Vainbrand, Prethvi Kashinkunti, J. Bernauer, Bryan Catanzaro, Amar Phanishayee, and M. Zaharia.
\newblock Efficient large-scale language model training on gpu clusters using megatron-lm.
\newblock \emph{International Conference for High Performance Computing, Networking, Storage and Analysis}, 2021.

\bibitem[Nichol and Dhariwal(2021)]{nichol2021improved}
Alexander~Quinn Nichol and Prafulla Dhariwal.
\newblock Improved denoising diffusion probabilistic models.
\newblock In \emph{ICML}, 2021.

\bibitem[Nichol et~al.(2022)Nichol, Dhariwal, Ramesh, Shyam, Mishkin, Mcgrew, Sutskever, and Chen]{nichol2022glide}
Alexander~Quinn Nichol, Prafulla Dhariwal, Aditya Ramesh, Pranav Shyam, Pamela Mishkin, Bob Mcgrew, Ilya Sutskever, and Mark Chen.
\newblock Glide: Towards photorealistic image generation and editing with text-guided diffusion models.
\newblock In \emph{ICML}, 2022.

\bibitem[Pan et~al.(2018)Pan, Lin, Fang, Huang, Zhou, and Lu]{pan2018recurrent}
Bowen Pan, Wuwei Lin, Xiaolin Fang, Chaoqin Huang, Bolei Zhou, and Cewu Lu.
\newblock Recurrent residual module for fast inference in videos.
\newblock In \emph{CVPR}, 2018.

\bibitem[Park et~al.(2019)Park, Liu, Wang, and Zhu]{park2019semantic}
Taesung Park, Ming-Yu Liu, Ting-Chun Wang, and Jun-Yan Zhu.
\newblock Semantic image synthesis with spatially-adaptive normalization.
\newblock In \emph{CVPR}, 2019.

\bibitem[Parmar et~al.(2022)Parmar, Zhang, and Zhu]{DBLP:conf/cvpr/Parmar0Z22}
Gaurav Parmar, Richard Zhang, and Jun{-}Yan Zhu.
\newblock On aliased resizing and surprising subtleties in {GAN} evaluation.
\newblock In \emph{{IEEE/CVF} Conference on Computer Vision and Pattern Recognition, {CVPR} 2022, New Orleans, LA, USA, June 18-24, 2022}, pages 11400--11410. {IEEE}, 2022.

\bibitem[Paszke et~al.(2019)Paszke, Gross, Massa, Lerer, Bradbury, Chanan, Killeen, Lin, Gimelshein, Antiga, et~al.]{paszke2019pytorch}
Adam Paszke, Sam Gross, Francisco Massa, Adam Lerer, James Bradbury, Gregory Chanan, Trevor Killeen, Zeming Lin, Natalia Gimelshein, Luca Antiga, et~al.
\newblock Pytorch: an imperative style, high-performance deep learning library.
\newblock In \emph{NeurIPS}, 2019.

\bibitem[Podell et~al.(2024)Podell, English, Lacey, Blattmann, Dockhorn, M{\"u}ller, Penna, and Rombach]{podell2023sdxl}
Dustin Podell, Zion English, Kyle Lacey, Andreas Blattmann, Tim Dockhorn, Jonas M{\"u}ller, Joe Penna, and Robin Rombach.
\newblock Sdxl: Improving latent diffusion models for high-resolution image synthesis.
\newblock In \emph{ICLR}, 2024.

\bibitem[Rajbhandari et~al.(2019)Rajbhandari, Rasley, Ruwase, and He]{rajbhandari2019zero}
Samyam Rajbhandari, Jeff Rasley, Olatunji Ruwase, and Yuxiong He.
\newblock Zero: Memory optimizations toward training trillion parameter models.
\newblock \emph{Sc20: International Conference For High Performance Computing, Networking, Storage And Analysis}, 2019.

\bibitem[Ramesh et~al.(2021)Ramesh, Pavlov, Goh, Gray, Voss, Radford, Chen, and Sutskever]{ramesh2021zero}
Aditya Ramesh, Mikhail Pavlov, Gabriel Goh, Scott Gray, Chelsea Voss, Alec Radford, Mark Chen, and Ilya Sutskever.
\newblock Zero-shot text-to-image generation.
\newblock In \emph{ICML}, 2021.

\bibitem[Ramesh et~al.(2022)Ramesh, Dhariwal, Nichol, Chu, and Chen]{ramesh2022hierarchical}
Aditya Ramesh, Prafulla Dhariwal, Alex Nichol, Casey Chu, and Mark Chen.
\newblock Hierarchical text-conditional image generation with clip latents.
\newblock \emph{arXiv preprint arXiv:2204.06125}, 2022.

\bibitem[Rasley et~al.(2020)Rasley, Rajbhandari, Ruwase, and He]{rasley2020deepspeed}
Jeff Rasley, Samyam Rajbhandari, Olatunji Ruwase, and Yuxiong He.
\newblock Deepspeed: System optimizations enable training deep learning models with over 100 billion parameters.
\newblock In \emph{Proceedings of the 26th ACM SIGKDD International Conference on Knowledge Discovery \& Data Mining}, pages 3505--3506, 2020.

\bibitem[Ren et~al.(2021)Ren, Rajbhandari, Aminabadi, Ruwase, Yang, Zhang, Li, and He]{DBLP:conf/usenix/0015RARYZ0H21}
Jie Ren, Samyam Rajbhandari, Reza~Yazdani Aminabadi, Olatunji Ruwase, Shuangyan Yang, Minjia Zhang, Dong Li, and Yuxiong He.
\newblock Zero-offload: Democratizing billion-scale model training.
\newblock In \emph{2021 {USENIX} Annual Technical Conference, {USENIX} {ATC} 2021, July 14-16, 2021}, pages 551--564. {USENIX} Association, 2021.

\bibitem[Ren et~al.(2018)Ren, Pokrovsky, Yang, and Urtasun]{ren2018sbnet}
Mengye Ren, Andrei Pokrovsky, Bin Yang, and Raquel Urtasun.
\newblock Sbnet: Sparse blocks network for fast inference.
\newblock In \emph{CVPR}, 2018.

\bibitem[Riegler et~al.(2017)Riegler, Osman~Ulusoy, and Geiger]{riegler2017octnet}
Gernot Riegler, Ali Osman~Ulusoy, and Andreas Geiger.
\newblock Octnet: Learning deep 3d representations at high resolutions.
\newblock In \emph{CVPR}, 2017.

\bibitem[Rombach et~al.(2022)Rombach, Blattmann, Lorenz, Esser, and Ommer]{rombach2022high}
Robin Rombach, Andreas Blattmann, Dominik Lorenz, Patrick Esser, and Bj{\"o}rn Ommer.
\newblock High-resolution image synthesis with latent diffusion models.
\newblock In \emph{CVPR}, 2022.

\bibitem[Ronneberger et~al.(2015)Ronneberger, Fischer, and Brox]{ronneberger2015u}
Olaf Ronneberger, Philipp Fischer, and Thomas Brox.
\newblock U-net: Convolutional networks for biomedical image segmentation.
\newblock In \emph{Medical Image Computing and Computer-Assisted Intervention--MICCAI 2015: 18th International Conference, Munich, Germany, October 5-9, 2015, Proceedings, Part III 18}, pages 234--241. Springer, 2015.

\bibitem[Saharia et~al.(2022)Saharia, Chan, Saxena, Li, Whang, Denton, Ghasemipour, Gontijo~Lopes, Karagol~Ayan, Salimans, et~al.]{saharia2022photorealistic}
Chitwan Saharia, William Chan, Saurabh Saxena, Lala Li, Jay Whang, Emily~L Denton, Kamyar Ghasemipour, Raphael Gontijo~Lopes, Burcu Karagol~Ayan, Tim Salimans, et~al.
\newblock Photorealistic text-to-image diffusion models with deep language understanding.
\newblock \emph{NeurIPS}, 2022.

\bibitem[Salimans and Ho(2021)]{salimans2021progressive}
Tim Salimans and Jonathan Ho.
\newblock Progressive distillation for fast sampling of diffusion models.
\newblock In \emph{ICLR}, 2021.

\bibitem[Shi and Chu(2017)]{shi2017speeding}
Shaohuai Shi and Xiaowen Chu.
\newblock Speeding up convolutional neural networks by exploiting the sparsity of rectifier units.
\newblock \emph{arXiv preprint arXiv:1704.07724}, 2017.

\bibitem[Shih et~al.(2023)Shih, Belkhale, Ermon, Sadigh, and Anari]{shih2023paradigms}
Andy Shih, Suneel Belkhale, Stefano Ermon, Dorsa Sadigh, and Nima Anari.
\newblock Parallel sampling of diffusion models.
\newblock \emph{NeurIPS}, 2023.

\bibitem[Sohl-Dickstein et~al.(2015)Sohl-Dickstein, Weiss, Maheswaranathan, and Ganguli]{sohl2015deep}
Jascha Sohl-Dickstein, Eric Weiss, Niru Maheswaranathan, and Surya Ganguli.
\newblock Deep unsupervised learning using nonequilibrium thermodynamics.
\newblock In \emph{ICML}, 2015.

\bibitem[Song et~al.(2020{\natexlab{a}})Song, Meng, and Ermon]{song2020denoising}
Jiaming Song, Chenlin Meng, and Stefano Ermon.
\newblock Denoising diffusion implicit models.
\newblock In \emph{ICLR}, 2020{\natexlab{a}}.

\bibitem[Song et~al.(2020{\natexlab{b}})Song, Sohl-Dickstein, Kingma, Kumar, Ermon, and Poole]{song2020score}
Yang Song, Jascha Sohl-Dickstein, Diederik~P Kingma, Abhishek Kumar, Stefano Ermon, and Ben Poole.
\newblock Score-based generative modeling through stochastic differential equations.
\newblock In \emph{ICLR}, 2020{\natexlab{b}}.

\bibitem[Song et~al.(2023)Song, Dhariwal, Chen, and Sutskever]{song2023consistency}
Yang Song, Prafulla Dhariwal, Mark Chen, and Ilya Sutskever.
\newblock Consistency models.
\newblock 2023.

\bibitem[Tang et~al.(2022)Tang, Liu, Li, Lin, and Han]{tang2022torchsparse}
Haotian Tang, Zhijian Liu, Xiuyu Li, Yujun Lin, and Song Han.
\newblock Torchsparse: Efficient point cloud inference engine.
\newblock In \emph{MLSys}, 2022.

\bibitem[Tang et~al.(2023)Tang, Yang, Liu, Hong, Yu, Li, Dai, Wang, and Han]{tangandyang2023torchsparse}
Haotian Tang, Shang Yang, Zhijian Liu, Ke Hong, Zhongming Yu, Xiuyu Li, Guohao Dai, Yu Wang, and Song Han.
\newblock Torchsparse++: Efficient training and inference framework for sparse convolution on gpus.
\newblock In \emph{MICRO}, 2023.

\bibitem[Vahdat et~al.(2021)Vahdat, Kreis, and Kautz]{vahdat2021score}
Arash Vahdat, Karsten Kreis, and Jan Kautz.
\newblock Score-based generative modeling in latent space.
\newblock 34:\penalty0 11287--11302, 2021.

\bibitem[Valiant(1990)]{valiant1990bridging}
Leslie~G. Valiant.
\newblock A bridging model for parallel computation.
\newblock \emph{Commun. ACM}, 33\penalty0 (8):\penalty0 103–111, 1990.

\bibitem[Wu and He(2018)]{wu2018group}
Yuxin Wu and Kaiming He.
\newblock Group normalization.
\newblock In \emph{ECCV}, 2018.

\bibitem[Xiao et~al.(2023)Xiao, Yin, Freeman, Durand, and Han]{xiao2023fastcomposer}
Guangxuan Xiao, Tianwei Yin, William~T. Freeman, Frédo Durand, and Song Han.
\newblock Fastcomposer: Tuning-free multi-subject image generation with localized attention.
\newblock \emph{arXiv}, 2023.

\bibitem[Xiao et~al.(2022)Xiao, Kreis, and Vahdat]{xiao2022DDGAN}
Zhisheng Xiao, Karsten Kreis, and Arash Vahdat.
\newblock Tackling the generative learning trilemma with denoising diffusion {GAN}s.
\newblock In \emph{ICLR}, 2022.

\bibitem[Xu et~al.(2021)Xu, Lee, Chen, Hechtman, Huang, Joshi, Krikun, Lepikhin, Ly, Maggioni, Pang, Shazeer, Wang, Wang, Wu, and Chen]{xu2021gspmd}
Yuanzhong Xu, HyoukJoong Lee, Dehao Chen, Blake Hechtman, Yanping Huang, Rahul Joshi, Maxim Krikun, Dmitry Lepikhin, Andy Ly, Marcello Maggioni, Ruoming Pang, Noam Shazeer, Shibo Wang, Tao Wang, Yonghui Wu, and Zhifeng Chen.
\newblock Gspmd: General and scalable parallelization for ml computation graphs.
\newblock \emph{arXiv preprint arXiv: 2105.04663}, 2021.

\bibitem[Yuan et~al.(2021)Yuan, Li, Cheng, Liu, Guo, Cai, Yao, Yang, Yi, Wu, Zhang, and Zhao]{yuan2021oneflow}
Jinhui Yuan, Xinqi Li, Cheng Cheng, Juncheng Liu, Ran Guo, Shenghang Cai, Chi Yao, Fei Yang, Xiaodong Yi, Chuan Wu, Haoran Zhang, and Jie Zhao.
\newblock Oneflow: Redesign the distributed deep learning framework from scratch.
\newblock \emph{arXiv preprint arXiv: 2110.15032}, 2021.

\bibitem[Zhang and Chen(2022)]{zhang2022fast}
Qinsheng Zhang and Yongxin Chen.
\newblock Fast sampling of diffusion models with exponential integrator.
\newblock In \emph{ICLR}, 2022.

\bibitem[Zhang et~al.(2022)Zhang, Tao, and Chen]{zhang2022gddim}
Qinsheng Zhang, Molei Tao, and Yongxin Chen.
\newblock gddim: Generalized denoising diffusion implicit models.
\newblock 2022.

\bibitem[Zhang et~al.(2023)Zhang, Song, Huang, Chen, and yu~Liu]{zhange2023diffcollage}
Qinsheng Zhang, Jiaming Song, Xun Huang, Yongxin Chen, and Ming yu Liu.
\newblock Diffcollage: Parallel generation of large content with diffusion models.
\newblock In \emph{CVPR}, 2023.

\bibitem[Zhang et~al.(2018)Zhang, Isola, Efros, Shechtman, and Wang]{zhang2018perceptual}
Richard Zhang, Phillip Isola, Alexei~A Efros, Eli Shechtman, and Oliver Wang.
\newblock The unreasonable effectiveness of deep features as a perceptual metric.
\newblock In \emph{CVPR}, 2018.

\bibitem[Zhao et~al.(2023)Zhao, Gu, Varma, Luo, Huang, Xu, Wright, Shojanazeri, Ott, Shleifer, et~al.]{zhao2023pytorch}
Yanli Zhao, Andrew Gu, Rohan Varma, Liang Luo, Chien-Chin Huang, Min Xu, Less Wright, Hamid Shojanazeri, Myle Ott, Sam Shleifer, et~al.
\newblock Pytorch fsdp: experiences on scaling fully sharded data parallel.
\newblock \emph{arXiv preprint arXiv:2304.11277}, 2023.

\bibitem[Zheng et~al.(2022)Zheng, Li, Zhang, Zhuang, Chen, Huang, Wang, Xu, Zhuo, Xing, et~al.]{zheng2022alpa}
Lianmin Zheng, Zhuohan Li, Hao Zhang, Yonghao Zhuang, Zhifeng Chen, Yanping Huang, Yida Wang, Yuanzhong Xu, Danyang Zhuo, Eric~P Xing, et~al.
\newblock Alpa: Automating inter-and $\{$Intra-Operator$\}$ parallelism for distributed deep learning.
\newblock In \emph{16th USENIX Symposium on Operating Systems Design and Implementation (OSDI 22)}, pages 559--578, 2022.

\end{thebibliography}
}


\end{document}